# A survey on automated detection and classification of acute leukemia and WBCs in microscopic blood cells

Mohammad Zolfaghari[1] · Hedieh Sajedi[2]

## Abstract

Leukemia (blood cancer) is an unusual spread of White Blood Cells or Leukocytes (WBCs) in the bone marrow and blood. Pathologists can diagnose leukemia by looking at a person's blood sample under a microscope. They identify and categorize leukemia by counting various blood cells and morphological features. This technique is time-consuming for the prediction of leukemia. The pathologist's professional skills and experiences may be affecting this procedure, too. In computer vision, traditional machine learning and deep learning techniques are practical roadmaps that increase the accuracy and speed in diagnosing and classifying medical images such as microscopic blood cells. This paper provides a comprehensive analysis of the detection and classification of acute leukemia and WBCs in the microscopic blood cells. First, we have divided the previous works into six categories based on the output of the models. Then, we describe various steps of detection and classification of acute leukemia and WBCs, including Data Augmentation, Preprocessing, Segmentation, Feature Extraction, Feature Selection (Reduction), Classification, and focus on classification step in the methods. Finally, we divide automated detection and classification of acute leukemia and WBCs into three categories, including traditional, Deep Neural Network (DNN), and mixture (traditional and DNN) methods based on the type of classifier in the classification step and analyze them. The results of this study show that in the diagnosis and classification of acute leukemia and WBCs, the Support Vector Machine (SVM) classifier in traditional machine learning models and Convolutional Neural Network (CNN) classifier in deep learning models have widely employed. The performance metrics of the models that use these classifiers compared to the others model are higher. We propose providing models in detecting and classify acute leukemia and WBCs that use a combination of SVM and CNN classifiers in their classification step to achieve optimum performance metrics.

## 1 Introduction

The bone marrow is responsible for producing blood cells. Undoubtedly, one of the essential organs of the body is the circulatory system. The task of the blood is to transfer oxygen and minerals to other parts of the body. It also keeps body temperature and protects the body through antibody production. Human blood comprises three main components Red Blood Cells (RBCs), WBCs and Platelets (Thrombocytes). RBCs carry oxygen to the body's tissues and collect waste materials from them. WBCs are a vital part of the human immune system for the defence of the body against disease and foreign invaders. Platelets are a group of blood cells that prevent bleeding [14]. There are two major groups of WBCs in terms of function: B and T lymphatic cells [9]. WBCs are several main categories, namely basophils, eosinophils, lymphocytes, monocytes, neutrophils, etc. [26, 40, 75]. Figure 1 shows the main types of WBCs. Blood components and the main types of WBCs are shown in Fig. 2.

Leukemia is a type of blood cancer that usually starts in the bone marrow and causes abnormal WBCs called blasts or leukemia cells. These cells cause stop normal WBCs and reduce the body's resistance [7]. Types of leukemia classify according to the kind of WBCs affected and according to the speed of disease progression. Acute leukemia begins suddenly and progresses rapidly over days or weeks. As a result, the patient's condition has worsened and should be promptly treated. Acute leukemia may lead to lethal death within a few months if not treated well. Chronic leukemia grows slowly and can take months or even years [8, 14]. There are different types of blood leukemia, and here, we describe four main types of them [38]:

1. Acute Lymphoblastic Leukemia (ALL) disease especially starts with B lymphoblasts. Its main symptoms are the increase of blast cells in the bone marrow and the decrease number of normal blood cells. The quick-growing of abnormal lymphocytes is called lymphoblast. When ALL happens that WBCs fully mature, this type of leukemia is more common in children. Lack of treatment of this disease causes many lymphoblasts are produced in the body and can lead to death [14, 51, 59]. ALL occurs more often in B than in T cells. For the first time, the French American British (FAB) performed the classification of ALL according to morphological criteria, and it divided into three subgroups: L1, L2, and L3 [12, 58, 62, 67].
2. Acute Myeloid Leukemia (AML) is a type of leukemia that occurs in bone marrow cells or Myelocytes and grows rapidly. Myelocytes make RBCs, WBCs (except granulocytes and lymphocytes), and platelets. The bone marrow in this disease produces myeloblasts, abnormal RBCs, or platelets that the most common malignant myeloid disorder in adults. AML is categorized into eight subtypes, including M0, M1, M2, M3, M4, M5, M6, and M7 [74].
3. Chronic Lymphoblastic Leukemia (CLL) is the most typical kind of leukemia that is more common in the elderly. The blood cells appear mature in this disease but are not

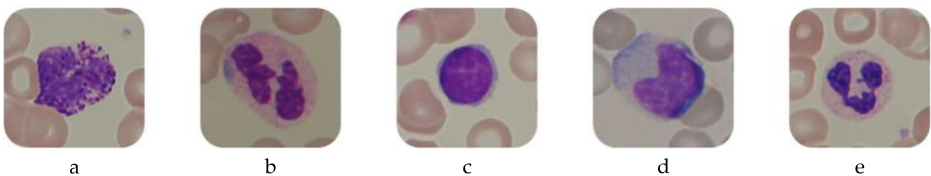

Fig. 1 Five main types of WBCs: (a) basophil, (b) eosinophil, (c) lymphocyte, (d) monocyte, (e) neutrophil

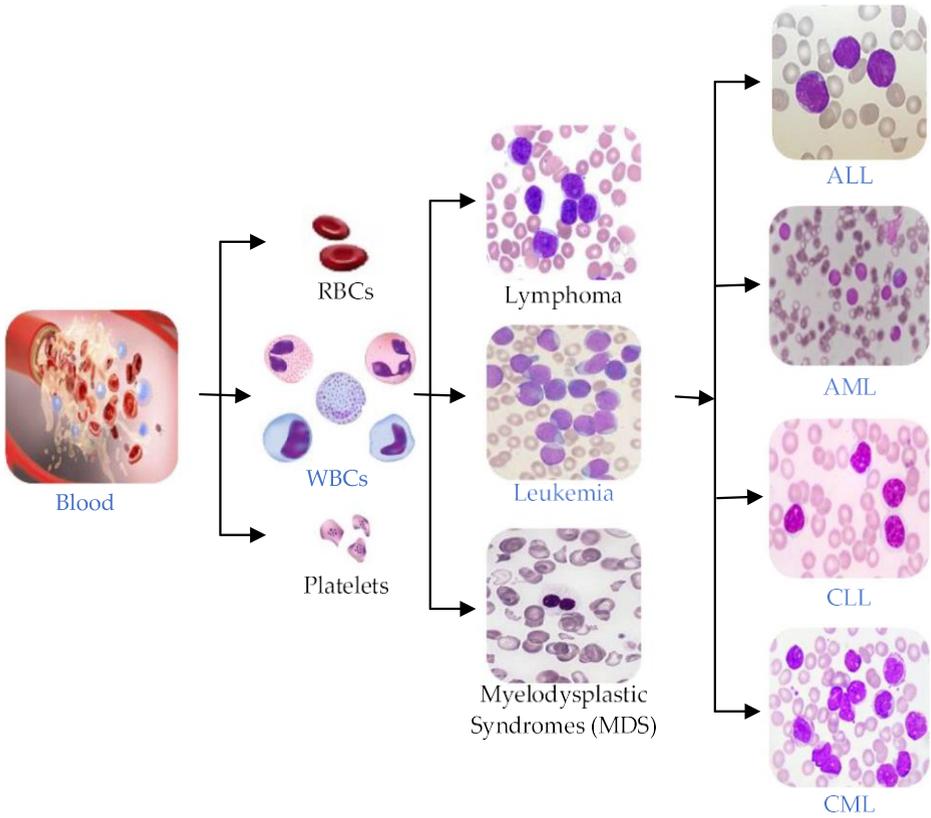

Fig. 2 Blood components and main types of leukemia

    completely normal and cannot fight off the invading cells. It may also spread to lymph nodes and organs such as the liver and spleen. CLL grows when too many abnormal lymphocytes grow, crowding out normal blood cells and weaken the immune system [14].
4. Chronic Myeloid Leukemia (CML) is a disease caused by an increase in the growth and concentration of bone marrow myeloid cells and an increase in their density in the blood. The progression of the disease is slow and controllable. CML patients have a normal life and are often asymptomatic [14].

The block diagram of the blood components and the main types of leukemia shown in Fig. 2.
  Pathologists and hematologists can diagnose leukemia by manually observing the patient's blood smear under a microscope. They use structural features and enumeration of cells to diagnose and classify leukemia and need advanced laboratory equipment to diagnose the type of leukemia. Diagnosis and classification of leukemia in this way is costly and time-consuming. Also, the accuracy of diagnosis and classification depends on the knowledge and experience of experts in this field. One of the methods that can solve these challenges and problems is using machine learning methods in computer vision. Using the traditional machine learning and deep learning method, the diagnosis and classification of acute leukemia and WBCs can be performed automatically, more easily and at a lower cost.

In this study, we provide a complete and comprehensive review of leukemia's automatic diagnosis and classification by studying the past literature of previous works in this field. First, we downloaded from the internet and about 114 resources, including journals, conferences, books, etc., about detecting and classifying acute leukemia and WBCs in microscopic blood cells and read them. Then, we have selected about 81 more important and relevant works from 2012 to 2020 that have shown in Fig. 3.

We categorize the previous works into the following six groups based on what type of acute leukemia classification performed or model output, as shown in Fig. 4:

1. Leukemia (General) and healthy
2. ALL and healthy
3. Subtypes of ALL (L1, L2 and L3) and healthy
4. AML and healthy
5. ALL, AML and healthy
6. Main types of WBCs

Machine Learning (ML) is one of the most prominent Artificial Intelligence (AI) fields now. It is the common method used in image classification. Image classification has constantly been a hot research topic. The output data label is available in supervised machine learning and we must use a supervised machine learning algorithm. Images in the WBC databases have the label. Therefore, acute leukemia and WBCs classification is a supervised problem. In supervised learning, traditional and deep learning algorithms commonly employed in the classification step of a classifier [49, 60, 81]. We divide previous studies in acute leukemia and WBCs classification into traditional, deep, and mixed models based on supervised classifier algorithms. We expect that the paper covers more methods of detection and classification of acute leukemia and WBCs. We will demonstrate:

- Computer-based methods have more used in acute leukemia and WBCs detection and classification, including Traditional and DNN models, in the last decade.
- As shown in Fig. 5, traditional and DNN methods are use more.

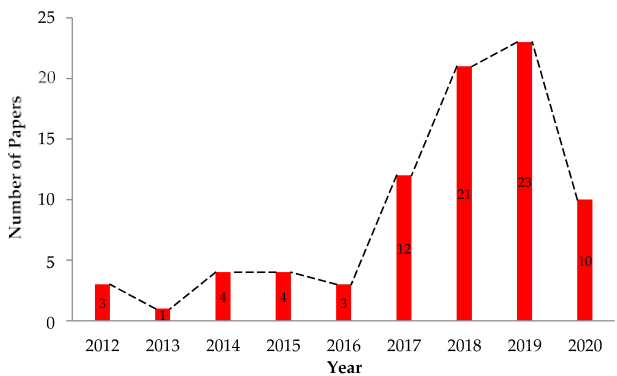

Fig. 3 Distribution researches for acute leukemia detection and classification by year

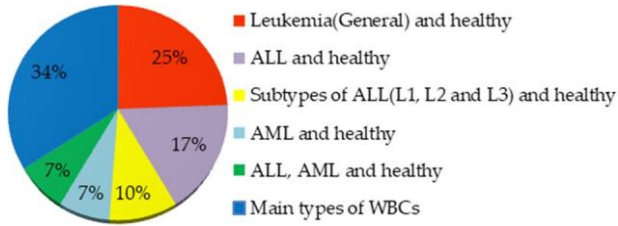

Fig. 4 The numbers of papers based on the type of acute leukemia classification

- The research results show that SVM among the traditional methods and Convolutional Neural Networks among the DNN methods classifier are more used and are achieved good results in most tasks.

The structural components of this paper described in the following: Previous studies in diagnosing and classifying acute leukemia and WBCs provided in Section 2. In Section 3, we review a common database for the detection and classification of ALL and classify previous studies that evaluate with this database into three categories traditional, deep, and hybrid and compared them in terms of accuracy. Methodology for acute leukemia and WBCs detection and classification discusses in Section 4. We compare performance metrics in previous works in Section 5. Finally, we discuss and conclude our study in Sections 6 and 7.

## 2 Previous studies in the field of diagnosis and classification of acute leukemia and WBCs

Generally, a machine learning system has two main parts, including feature extraction and classification. The feature extraction and classification steps in traditional learning perform separately. Deep learning is a method to implement many machine learning algorithms using multi-layers neural networks. These multiple processing layers learn demonstrations of data with multiple levels of abstraction for considering the input data. In deep learning, feature extraction and classification steps achieve in a single shot that can automatically learn features from data.

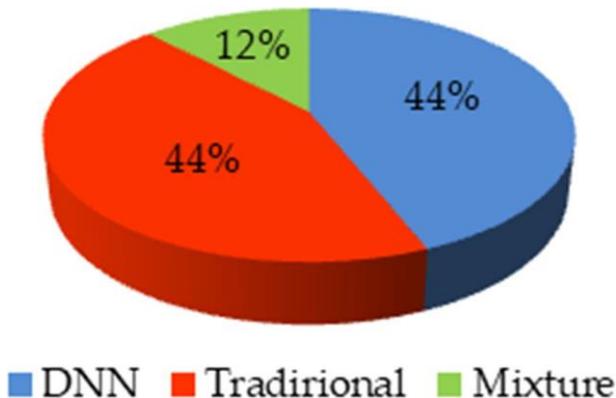

Fig. 5 The numbers of papers based on the type of employed network method

In the paper, we perform a comprehensive and complete study in diagnosing and classifying acute blood leukemia and WBCs. Then we divide the previous works into six groups according to the type of output of the models (see section 1):

## 2.1 Leukemia (general) and healthy

Patel et al. proposed a method for the detection of leukemia from blood microscopic images [52]. They applied histogram equalization and Zack algorithm for categorizing WBCs and used K-means clustering for WBCs detection. The model evaluated and achieved 93.57% accuracy with SVM. The main focus of this work is to suggest an automatic model that identifies leukemia from microscopic blood images and decreases the time of detection than the manual method.

Ravikumar introduced a fast Relevance Vector Machine (RVM) for WBC image segmentation [56]. Fast RVM is a modification of RVM that has a much faster testing time. They also used Single Hidden Layer Feed Forward Neural Networks (SLFN) for classification. The algorithm is a version of Extreme Learning Machine (ELM) with a higher learning rate and generalization. They compared the performance metrics of the models. The fast RVM model was more accurate and had less execution time than other models. The method was effective for WBC identification and efficiently decreased the effects of illumination and staining. Therefore, this approach had robust flexibility and high computational efficiency.

Vogado et al. presented a leukemia detecting scheme by transfer learning [71]. CNN (i.e. AlexNet [34], CaffeNet, and Vgg-f) were used for the feature extraction and employed gain ratio for the feature selection step. Classification was done by CNN and SVM classifiers on 377 images. Three various datasets were applied for validation, and classification accuracy of 99.2% was attained. The key advantage of the method removes the segmentation step because using CNN was used for the automatic feature extraction step.

Patil et al. designed ALL discovery consist of preprocessing, feature extracting, feature selection, and classification for discriminate normal from abnormal WBCs [53]. Discrete Orthonormal S-Transform (DOST) is a multi-resolution method for the categorized texture of an image. They used DOST for feature extraction and Principal Component Analyzes (PCA) [54], and Linear Discriminant Analyzes (LDA) [19] for feature reduction. Classification is done with AdaBoost based Random Forest (RF) classifier. They helped in improving the performance of the model with feature extraction from the nucleus and cytoplasm of WBCs.

Sahlol et al. suggested a method to detect WBCs Leukemia [61]. Features of WBCs were extracted by DNN (VGG-19), and feature selection made by Statically Enhanced Slap Swarm Algorithm (SESSA). They applied SVM in the classification step. The model was evaluated on a database with 260 images and attained 96.11% classification accuracy. They tested the model and similar methods on the ALL-IDB2 and C-NMC database and compared them. The results show that the method is better performance with fewer resource consumption and effective use of storage capacity than related works.

Gayathri and Jyothi produced two methods for WBCs classification [21]. The first method was based on traditional feature extraction. In traditional feature extraction, the feature like the area, major and minor axis, number of the nucleus are extracted then are forwarded to SVM and Artificial Neural Network (ANN) classifier for classification. They used Adaptive K-Means Clustering (AKM) for better segmentation. The second method microscopic leukocyte exposed to CNN. These models trained with 47 images and tested with 35 images of the database. The accuracy of the first model with the SVM classifier was 89.47% and with ANN was 92.10%. The Accuracy of the model with CNN was 93%. We conclude that using AKM in the segmentation step for extracting the nucleus in the model helps to easier image for a system to further process and classification accuracy with CNN is better than ANN and SVM.

Loey et al. produced two automatic classification methods for classifying normal and abnormal microscopic blood cells [42]. Image preprocessing was made by translated, reflected, and rotated images. The first method (AlexNet) used for feature extraction and classification employed with Linear Discriminants (LDs) and K-Nearest Neighbors (K-NN). In the second method, AlexNet applied for feature extraction and classification steps. Two methods evaluated with 2820 images. The average accuracy of the first method was 98.69% and the second method 100%. The authors' purpose was to compare the classical (first model) performance metrics and deep (second model) models in diagnosing and classifying acute leukemia. Comparing the results of the two models, we see that the deep model, which is based on AlexNet, performed better than the traditional model in all performance metrics.

Vogado et al. presented a computational model for leukemia detection [72]. They called the model LeukNet. LeukNet was a constructed CNN based VGG-16 convolutional blocks. They evaluated LeukNet with 3536 WBC Images and achieved a classification accuracy of %98.20. Architecture, parameters and fine-tuning system in the model cause to improve a model for diagnosis that is more accurate and robust.

Vogado et al. developed a model for leukemia recognition [70]. First, the model use VGG-f, a type of CNN, for feature extraction. Then, the feature vector is reduced by PCA. They created an Ensemble of Classifiers (EOC) with three classifiers, including SVM, Multi-Layer Perceptron (MLP), and RF for categorized normal and abnormal WBCs. The model was evaluated with 108 WBCs images and reached 100%classification accuracy. They compared AlexNet, CoffeNet, Vgg-f and their mixtures in terms of feature vector size and accuracy. The results showed that the features coming from Vgg-f are the ones that present the maximum gain ratio. Feature extraction with vgg-f has a smaller computational time in comparison to the combination of architectures. So, they used this type of feature extraction step and EOC in the classification step for achieving high accuracy. It seems that these two steps were very important in model accuracy.

Liu and Long developed an automatic system for ALL detection and classification [41]. They employed enhanced bagging ensemble training to classify health cells (denote as HEM) and ALL cells. The model involved two Inception ResNets that are called model-A and model-B. First, pre-training is done on the training set A and B. Then, the model trained in the previous step combined. The mixtures made the classification of the A and B models and joint ones. The model achieved 90% classification accuracy after 20 epochs. The lack of training samples and minor visual difference between ALL and healthy cells are two challenges in acute leukemia detection and classification. The method introduced overcoming the two challenges. Data augmentation used for increasing database samples and employed transfer learning with Inception and ResNets architecture for the minor visual difference between ALL and healthy cells.

2.2 ALL and healthy

Mohapatra and Patra presented a model-based EOC for detection ALL from normal blood cells [47]. The EOC included Naive Bayes (NB), K-NN, MLP, Radial Basis Functional Network (RBFN), and SVM. They employed Shadowed C-Means clustering (SCM) in the segmentation step and forty-four feature extracted. The EOC applied in the classification step.

The model evaluated with 5-fold cross-validation and achieved 96.88% accuracy. The approach was using EOC in the classification step.

Abdeldaim et al. presented a new traditional method to category normal from abnormal cells AML of WBCs [1]. First, they used a mixture of Zack algorithm and histogram equalizer for WBCs segmentation. Then, color, shape, and texture were extracted in the feature extraction step. Finally, WBCs are categorized by K-NN, SVM, NB, and Decision Tree (DT) classifiers. The best classification accuracy with a dataset including 260 images was 96.01% by the K-NN classifier. The metod almost was similar to Mohapatra model, and the accuracy of the classification was near to that.

Hariprasath et al. used blast cell morphological features to detect and classify ALL [25]. Image segmentation applied by histogram equalization, Zack thresholding (triangle method) [76], dilation, erosion, and selects the nucleus and obtains cytoplasm from the cell. Statistical and texture features were extracted in the feature extraction step, and classification was conducted with SVM and K-NN. The model was tested on two databases (without noise and noisy). The test results of the model showed that the K-NN on the noise-free database and SVM on the noisy database is more effective. The model accuracy with K-NN was 91.33% and with SVM was 91.5%.

Thanh et al. presented a shallow CNN with seven layers for distinguishing between normal and abnormal blood cells images [68]. The first five layers performed feature extraction, and the next two layers performed classification. The database was small, and some simple data augmentation methods such as blurring, histogram equalization, reflection, translation, rotation, shearing applied to increase the data. The database includes 108 cell images and the increase to 1188 images after the data augmentation step. Accuracy of the classifier in the best case was 96.6%. The approach focuses on the data augmentation technique for DNN. In other words, aiming to prevent overfitting and increase efficiency in the model.

Ahmed et al. introduced a CNN model with two convolution layers, two max-pooling layers, and fully connected layers with 128 neurons for detection and classification of four subtypes of leukemia [4]. They used the data augmentation step to increase samples of the database and prevent overfitting. Five output nodes were organized by SoftMax activation function for the classification of four subtypes of leukemia and normal WBCs. The classification accuracy of the model was 88.25% in the best case. The work was similar to the Thanh model in architecture and data enhancement techniques to prevent overfitting but had lower performance metrics than the Thanh model.

Bodzas et al. suggested a new method for automatic classification ALL [15]. They applied new segmentation steps with two phases for the extraction of robust features. The classification step was done with SVM and ANN classifier. The model evaluated on 241 images and achieved 96.72% classification accuracy with SVM and 97.52% with ANN. The approach used a new technique in the feature extraction step for excruciating nucleus from cytoplasm features. Like some methods, they employed ANN and SVM in the classification step and comparing of them, too.

## 2.3 Subtypes of ALL (L1, L2 and L3) and healthy

Mirmohammadi et al. introduced a method for identifying subtypes of ALL based multi SVM classifier [45]. After Image Enhancement, They segmented images with nuclei segmentation. K-means algorithm used to segmentation nuclei in a similar approach, but the algorithm gains a blank cluster. They employed Fuzzy C-Means clustering (FCM) [77] to avoid empty cluster.

PCA is employed in the feature extraction step to improve accuracy. The model accuracy was 97%.

Rehman et al. suggested an automatic classification method for ALL and ALL subtypes [48]. Furthermore, they proposed a simple segmentation method based on a simple threshold method and transfer learning that was new methods. WBCs were detected using a simple threshold method and classify normal WBCs and subtypes of ALL with CNN based on AlexNet with seven layers (five layers of convolution and max-pooling and two layers fully connected layers). The experiment results on 330 images of the dataset were 97.78% classification accuracy.

Shafique et al. suggested an automatic system to detect and classify normal and subtypes of ALL from WBCs [62]. Data augmentation step is applied with Deep CNN (DCNN) to prevent overfitting. The model was based on AlexNet with five convolution and pooling layers and four fully connected layers for classification L1, L2, and L3. The model was evaluated on the dataset with 760 WBCs images after data augmentation and achieved a classification accuracy of 99.5%. The method has effective, and also it is better than the prior standard approaches since it doesn't require segmentation and performs automatically with DCNN.

### 2.4 AML and healthy

Wiharto et al. proposed a method that classified M0 and M1 cells in AML [74]. Dataset includes 50 images. The segmentation step was used through Otsu's Thresholding [48], and three important morphological features were extracted in the classification step. Synthetic Minority Oversampling Technique (SMOTE) applied with RF to overcome the imbalance of data. The classification accuracy of the model was 89.6%. They used SMOTE with RF classifier for avoiding the imbalance database.

Dasariraju et al. suggested a method for identification and categorization AML with RF for AML [17]. WBCs segmentation was made with multi Otsu's Thresholding and morphological operations. Sixteen feature of each image of WBCs extracted, and the five most important features used in the classification step. The number of images from each kind of leukocyte was 1274. The method achieved 93.45% detection accuracy and 93.45% classification accuracy. The model achieved a good detection and classification using Gini importance in the feature extraction step and used the five most important of them in the classification step.

### 2.5 ALL, AML and healthy

Supardi et al. presented a model for separates ALL and AML [66]. Twelve features were selected in the feature extraction step, and K-NN is employed in the classification step. They used four types of distance metric, including Euclidean, City block, Cosine and Correlation. K values from one until ten were tested. The optimal K value was four. The model tested with K = 4 and four types of distance metric on the 1500 images and achieved 86% average accuracy in the best case with Cosine distance. The authors of the paper say that the K-NN, can achieve good results for acute leukemia classification. But the accuracy of their method does not confirm such a statement.

Rawat et al. introduced a computer-assisted classification technique for the prediction of ALL and AML [57]. They segmented the nucleus from the leukocyte cell background on 420 images with Otsu's method. They evaluated 331 features of each segmented nucleus using a GA-SVM (Genetic Algorithm-SVM) and Radial Basis Kernel (RBK) in the classification step.

The model achieved 99.5% classification accuracy. They classified into various classes of AMC, ALC, and their FAB subtypes. Before, no such study was present in which leukocytes were classified into various ALL, AML, and subtypes of ALL and applied kernel methods and Genetic algorithm for ALL, AML and ALL subtypes. Therefore, the method is a new technique both in terms of classification method and type of classification.

Claro et al. presented a CNN model (Alert Net with a Residual Layer (Alert Net-R) and Alert Net with Depth wise Discrete Convolutions Layer (Alert Net-X)) to detection and classification ALL and AML [16]. The model evaluated with 2415 images on the sixteen datasets, and its accuracy was 97.18%. Diversity in the training data leads to the attainment of a strong system for different input image types. They used multi datasets in the training phase for real environment simulation. Also, the method has a smaller file size, which can be used to apply mobile systems.

## 2.6 Main types of WBCs

Gautam et al. suggested NB classifier and morphological features to classify the main five types of WBCs. [20] Otsu's Thresholding used to WBCs Segmentation and mathematical morphing to remove all structures that do n't similar WBCs. In the feature extraction, only the nucleus area was considered. The method was able to make 80.88% accuracy with 88 images. Using Otsu's Thresholding caused the model be very fast and provides accurate results in terms of good separation between foreground and background in the segmentation step.

Benomar et al. introduced multi features based method for WBCs segmentation and classification [13]. In the preprocessing step, a new color transformation and marker controlled watershed algorithm are employed. In the feature extraction step, a set of color, texture and morphological features are extracted to identify the nucleus from cytoplasm regions. RF is applied in the classification step, and model gets 95.86% classification accuracy. The benefit of the method over previous similar tasks is decreasing adjacency between cells (WBCs and other cells of blood). In the preprocessing step, the background and the RBCs surrounding the WBCs are removed from the image. Damaged cells and the false positive objects are cleaned with binary mask in this step, too.

A new system for WBCs identification based on CNN, named WBCsNet [64], was proposed by Shahin et al. Deep activation features and fine-tuning of existing deep networks are two methods based on transfer learning used in the approach. Several pre-trained networks are extracted by deep activation features and used in WBCsNet. They employed SVM in the classification step. The model tested on three different public WBCs datasets, including 2551 images, and achieved 96.1% classification accuracy. Using CNN in the segmentation step can have a positive influence on the overall system accuracy, and the model can be used as a pre-trained network in subtypes of WBCs identification and classification tasks.

Yu et al. developed an automatic cell detection method with deep learning for the classification of six types of leukocytes [75]. The model was based on CNN. They employed majority voting with ResNet50, Inception V3, VGG-16, VGG-19, and Xception in the classification step. The average accuracy in the WBCs classification was 88.5%. The model used several deep methods in the classification step, making the model with a relatively small database to achieve good accuracy. Experiments have also shown that it can also attain good results with blur images. The method gets good results with blur images. Therefore, it is very robust and efficient against large invariance forms of distortion in the input.

In another research, CNN was employed for the identification and categorization of five subtypes of WBCs [80]. Features extraction step included two parts. First, features of basophil and eosinophil are extracted and category with SVM. Then, a CNN used to automatic feature extraction of neutrophil, lymphocyte and monocyte, and RF classifiers applied in the classification step. The classification accuracy of the model was 92.8%. The method can identify WBC using the location of the nucleus of WBC, which is timeless and accurate.

A cataloguing method for five types of WBCs with transfer learning and deep learning was proposed Habibzadeh et al. [24]. They tried to consider preprocessing and supervised classification of four major types of WBCs with a modified deep learning architecture. After the preprocessing step, Features are extracted with transfer learning. Inception and ResNet applied in the classification step. Evaluation of the model is made with 1244 images of WBC and attained 99.84% accuracy. The approach has high performance for detecting and classify the main types of WBC by combining and modifying a small part of the popular CNN methods.

Lin et al. suggested an approach for diagnosis subtypes of WBCs [40]. They improved K-means clustering, Feature Weight Adaptive K-Means Clustering (FWSA-KM clustering algorithm), extracting complex WBCs and applied CNN in the classification step. The model was evaluated with 368 images. Maximum classification accuracy was 98.96%. One of the advantages of using CNN in the classification of WBCs is that there is no need for much preprocessing. We know this advantage as the reason for using the CNN architecture in the model.

Macawile et al. suggested a method that applied AlexNet, ResNet-101, and GoogleNet for classification main types of WBCs [24]. The model trains with 260 images on a dataset and get the best accuracy with AlexNet 96.63%. The approach with blob analysis for segmentation step and merge CNN in the classification step achieved more accurate results.

Wang et al. suggested PatternNet fused Ensemble of CNN (PECNN) to WBCs classification [73]. First, the number of CNN are created to structure an ensemble of the classifier. Then, the fusion mechanism is applied with several fusion algorithms. The algorithm adapted to data with randomly generated CNN forms. The model was executed with 300 images on the clean and noisy dataset. The average classification accuracy in the best case on the clean dataset was 99.90% and on the noisy dataset was 99.37%. So, the model will be robust with noisy datasets. The model used to ensemble classifiers for getting the best performance metrics. It has much less computational cost and relatively lightweight that can be used in mobile applications.

Banik et al. developed a CNN model for WBC classification [11]. They employed the model with ten layers (five convolutional layers, three max-pooling layers, one combined max-pooling layer and a fully connected layer). They used two pooling layer (max-pooling-11 and max-pooling-12) after the first convolutional layer and one max-pooling layer (max-pooling-51) after the fifth convolutional layer. Combined max-pooling layer (max-pooling-11 and max-pooling-12) created after max-pooling-12. They connected max-pooling-12 and combined max-pooling for the fusion of features, and the output of combined max-pooling gived to a fully connected layer. Dropout applied to avoid overfitting. The classification accuracy of the model was 98.61%. The model has higher speed and accuracy than similar models, such as the CNN-RNN in the Liang model.

Banik et al. proposed a new CNN method and nucleus segmentation for WBCs classification by fusing the first and last convolutional layers [11]. They used the dropout technique for preventing the model from overfitting problem. The model evaluated on four public datasets, including Blood Cell Count and Detection (BCCD) database, ALL-IDB2 database, JTSC database, CellaVision database and achieved 98.6% average accuracy and more than 97% on each database. Also, they tested the method on BCCD database with nine classification metrics and attained an overall accuracy 96%. The segmentation step in this approach increased generalization capability. Using WBCs localization caused reducing the performance dependency between the nucleus segmentation and classification technique.

An effectual segmentation step with a fully CNN for WBCs and RBCs classification introduced by Razzak et al. [29]. ELM and CNN are employed in the classification step. ELM classified WBC cells, and CNN categorized RBCs. The experiment was accompanied by 64,000 blood cells of the dataset. The average accuracy was 95.10%. Performance metrics of the model indicated that segmentation results on multi datasets are fast, robust, effective and coherent.

Liang et al. created a combining method with CNN and RNN (Recurrent Neural Networks), including bidirectional LSTM (Long Short Term Memory networks) to CNN-RNN [39]. They proposed a model with CNN-RNN and transfer learning for classification images of blood cells. The model involves a Pre-trained CNN layer, RNN layer, Merge layer and fully connected layer with dense Softmax output. It trained and evaluated with ResNet-50, Xception, InceptionV3, ResNet50-LSTM, Xception-LSTM, InceptionV3-LSTM, Xception-ResNet50-LSTM and the best accuracy was 90.79% with Xception-LSTM. Xception and RNN models used together for training large medical database. The model adaptively learned the characteristics of blood images and no need the segmentation step. They applied transfer learning to enhance joined model strength and quickness the convergence of the model in the classification step.

Hegde et al. suggested an automated model to discover five types of WBCs and classified them as leukemic and Non-leukemic [26]. They achieved 1159 images with different intensity and color shadows from Leishman stained marginal blood smears. They used three classifiers include SVM1, SVM2 and Neural Network for classification. SVM1 detected basophil cells. Lymphocyte, monocyte, neutrophil, and eosinophil cells were detected by the NN classifier. SVM2 was used to classified normal from abnormal WBCs in the classification step. Overall classification accuracy with 1159 images by combination SVM and NN achieved 98.8%. A robust system employed for calculating accurate threshold value for finding a region of cores, and a new method for mask identification of WBC from the cropped sub-images was implemented.

## 3 Acute lymphoblastic leukemia-image Data Base (ALL-IDB)

Detection of leukemia disease through microscopic medical blood image initiates with the peripheral blood images. There are many datasets for this work that can be helpful to researchers for testing and evaluating the models in image processing tasks.

ALL-IDB is one of the most common and free databases used for diagnosing, segmenting and classifying acute leukemia [36]. The ALL-IDB has two distinct kinds, including ALL-IDB1 and ALL-IDB2. This dataset includes about 39,000 blood cells and created in September 2005. ALL-IDB1 has 108 images (59 healthy images and 49 cancer images) containing multiple cores per image, and it used for segmentation and detection of WBCs. ALL-IDB2 contain 260 images that every image has a single lymphoblast per image and designed for evaluating the efficiency of classification methods. We have shown images sample of ALL-IDB1 (healthy and leukemic cell) in Fig. 6, and images sample of ALL-IDB2 (healthy and leukemic cell) shown in Fig. 7.

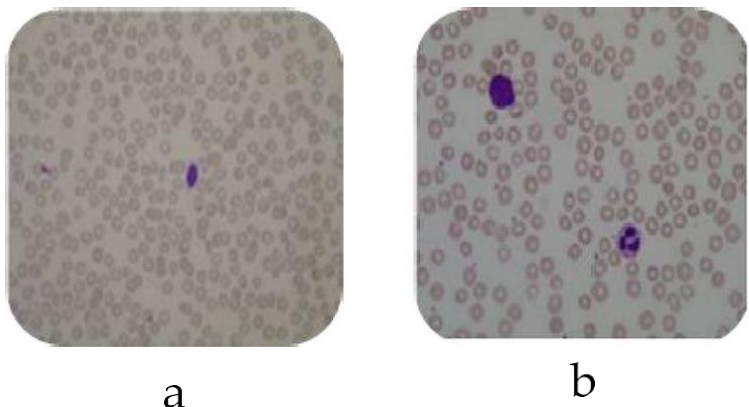

Fig. 6 Images sample of ALL-IDB1 (Healthy cell (a) and Leukemic cell (b))

Many images processing works have been performed with the ALL-IDB database in recent years. We have arranged some of the studies conducted on this database by database employed (ALL-IDB, ALL-IDB1, and ALL-IDB2), type of classification employed (traditional, DNN and mixture), year and the model accuracy in Table 1. Table 1 presents the previous study in the diagnosis and classification of leukemia that has used the ALL-IDB database in order of classification and year. Our purpose in presenting this table is to introduce the ALL-IDB database as small and a good database in Acute Leukemia detection and classification task. As we can see all the detection and classification methods obtain very good accuracy with three parts of the ALL-IDB database.

## 4 Methodology for acute leukemia and WBCs detection and classification

The basic structure of a machine learning system for the diagnosis and classification of acute leukemia and WBCs include the feature extraction and classification steps. Of course, the steps in Fig. 8 are based on extensive reviews of previous studies in this field, and the researcher may not do some steps in his/her work and achieve good results. All the above steps maybe do not need to be taken to achieve high-performance metrics. Therefore, we have described all the steps in this figure so that the reader understands the different steps in the diagnosis and classification of acute leukemia and obtains complete and comprehensive information from them. The different steps of detection and classification of acute leukemia are as follows (See Fig. 8): Data Augmentation, Preprocessing, Segmentation, Feature Extraction, Feature Selection (Reduction) and Classification.

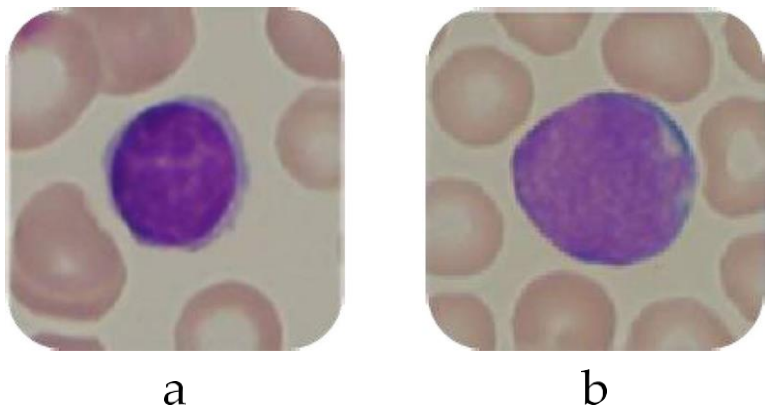

Fig. 7 Images sample of ALL-IDB2 ((a) Healthy cell and (b) Leukemic cell)

Table 1 ALL-IDB database works in the computer-based diagnosis and classification of leukemia

| Database employed | Work | Classification method | Year | Accuracy (%) |
|---|---|---|---|---|
| ALL-IDB | [52] | Traditional | 2015 | 93.57 |
| | [70] | Traditional | 2017 | 100 |
| | [32] | Traditional | 2019 | 99 |
| | [25] | Traditional | 2019 | 91.5 |
| | [80] | DNN | 2017 | 92.8 |
| | [43] | DNN | 2018 | 96.63 |
| | [58] | DNN | 2018 | 97.78 |
| | [40] | DNN | 2018 | 98.96 |
| | [4] | DNN | 2019 | 88.25 |
| | [2] | DNN | 2019 | 98.6 |
| | [29] | Mixture | 2017 | 95.1 |
| ALL-IDB1 | [55] | Traditional | 2013 | 92 |
| | [3] | Traditional | 2018 | 94 |
| | [46] | Traditional | 2019 | 99.66 |
| | [21] | DNN | 2018 | 93 |
| | [68] | DNN | 2018 | 96.6 |
| | [5] | DNN | 2019 | 97.07 |
| ALL-IDB2 | [65] | Traditional | 2014 | 88.79 |
| | [1] | Traditional | 2018 | 96.01 |
| | [69] | Traditional | 2018 | 96.25 |
| | [62] | DNN | 2018 | 96.06 |
| | [30] | DNN | 2019 | 98.7 |

Performance metrics of a machine learning system are important, but other metrics such as model complexity and execution time are also important. We examined models with all of the above steps, but the model was computationally challenging in terms of complexity and execution time.

In the following of this section, we describe the machine learning steps for diagnosing and classifying acute leukemia:

### 4.1 Data augmentation

Data augmentation step in image processing are methods employed to increase the number of images by addition somewhat changed fakes of already existing images or newly generated artificial images from existing images. Many techniques can be used to create artificial images according to the original images in the database. Some of them are as follows: Histogram equalization, Reflection and Shifting, Vertical and horizontal, Rotation, Shearing, Flipping, Zooming, etc.

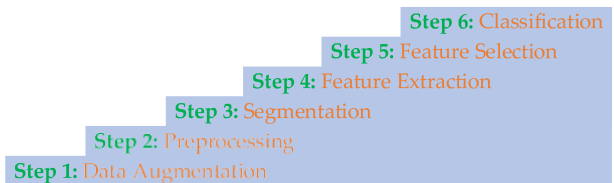

Fig. 8 Machine learning steps detection and classification of acute leukemia and WBCs

These methods are generally employed for decreasing and preventing overfitting, especially in the deep learning models [14]. The data augmentation step reduces the difference between training and test data [72]. In [4, 24, 27, 35, 42, 62, 68, 72, 73], data augmentation step was performed before starting network.

### 4.2 Image preprocessing

Image preprocessing is a technique of altering images suitable and quality of image improved for next steps. Different reasons can affect the quality of microscopic images. Researches suggested many techniques to identify and make the blood images suitable for segmenting the Region of Interest (ROI). They have enhanced the blood image by changing to another domain, such as convert RGB (R: Red, G: Green, B: Blue) to HSV (H: Hue, S: Saturation, V: Value) or to CMYK (Cyan, Magenta, Yellow, and Key (black)) domain to highlight the features of objects for efficiently detecting ROI. Many techniques such as Histogram equalization, Linear contrast stretching, Median filter, Minimum filter, Gaussian filter, Unsharp masking, and normalization are used to enhancing image quality [6] [63]. The input image to the network was RGB image color space in most of the tasks we examined.

In the preprocessing step of some works, the RGB image was changed to L*a*b color space as the input image of the network [59] [17] [22] [32] [2].L*a*b* is denoted by lightness (L*) and two color components are represented by (a*) and (b*). In [58] [28] transformed RGB to HSV color space as the input image format. Abdeldaim et al. converted RGB color space to CMYK as input for their model [1].

### 4.3 Image segmentation

Segmentation is an input image into the foreground and background region. Image is divided into different components or objects in this step. This division should stop when the objects or ROI have been insulated. Here, ROI can be WBCs in the microscopic image of blood [8]. Image segmentation includes many techniques such as thresholding (Otsu's Thresholding, Zack's Algorithm and maximum entropy method), Region-based segmentation (Region growing, Region splitting and merging [23, 78])), Watershed segmentation, Clustering-based segmentation(K-Means clustering [18], FCM, K-Medoids [31], Clustering based on Rough sets [79], Rough fuzzy C-means (RFCM) [37]), supervised segmentation approaches, frequency-domain methods, morphology segmentation (erosion, dilation, opening, and closing), Neural Network based methods (SVM, Functional link ANN (FLANN), CNN, etc.), segmentation by Active contour [33] etc. One of the most important and challenging steps in diagnosing and classifying acute leukemia is the segmentation step because the efficiency of the two next steps depends completely on this step. The taxonomy of WBCs segmentation methods for Acute Leukemia and WBCs are shown in Fig. 9.

As you can see in Table 2, most of the traditional methods we examined used the thresholding method in the segmentation step [1, 17, 20, 57]. In most DNN methods, the segmentation step is not used (See Table 3).

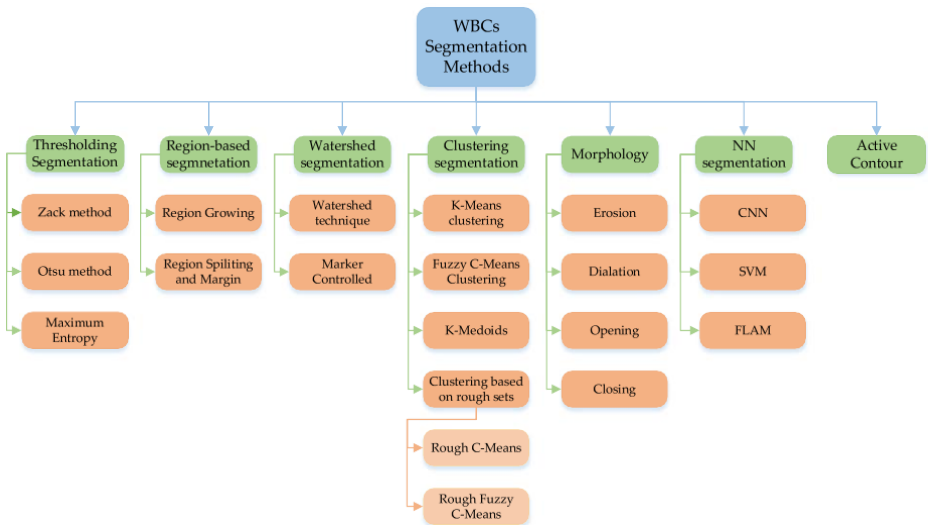

Fig. 9 WBCs segmentation methods for Acute Leukemia and WBCs

### 4.4 Feature extraction

The feature extraction step is applied to elicit and detect features derived from the segmented area or whole image. This section uses the texture or shape properties of the segmentation step and reduces the size of the image by removing redundant information from the original image. Therefore, it increases the processing speed and decreases execute time at this step [15]. Many features can be extracted from the objects in the image, such as the shape features, Texture Features, Statistical Features, Geometrical features, color features, etc. In deep learning networks, feature extraction is done automatically by networks [6]. Of course, traditional methods automatically performed feature extraction by DNN and achieved high accuracy [61, 71]. The taxonomy of WBCs feature extraction methods for Acute Leukemia and WBCs is shown in Fig. 10.

According to Table 2, we see that in the feature selection step, most traditional methods have used all three feature extraction methods, including color (color histogram), texture and shape feature. In [71], feature extraction is performed by the Deep method, and the accuracy of the model is 99.2%. The interesting point of this model is the use of SVM in the classification step, which is another reason for increasing the model's accuracy. A reearch has used a special CNN (VGG-19) to extract the feature, and the accuracy of this model is 96.11% [61]. The classifier of this model is also SVM. But in DNN, the feature extraction step is done automatically by these networks. In the feature extraction column of Table 3, we see the feature extraction step is performed using the DNN.

### 4.5 Feature selection

In the feature reduction step, essential and related features are recognized, and the vector dimensions attained from the prior step are reduced [53]. This step eliminates redundant features and reduces computational complexity and cost to help find accurate models [2].

Table 2  Traditional classification methods of leukemia based works

| Work | Year | Segmentation method | Feature Extraction method | Classifier(s) employed | Accuracy (%) | Remarks |
|---|---|---|---|---|---|---|
| | | | | **Leukemia (general)-Healthy** | | |
| [52] | 2015 | K- Mean clustering and Zack algorithm | Shape, color, statistical, and texture features | SVM | 93.57 | • Higher performance compared to the manual method |
| [56] | 2015 | With ELM, RVM, and fast-RVM | N/A | RVM and Fast-RVM | N/A | • Robust flexibility<br>• high computational efficiency |
| [71] | 2018 | Without segmentation | Deep features | SVM | 99.2 | • Using CNN in the feature extraction step |
| [53] | 2020 | Watershed | texture features | AdaBoost based RF | N/A | • Using nucleus and cytoplasm of WBCs in the feature extraction step |
| [61] | 2020 | – | VGGNet-19 | SVM | 96.11 | • Fewer resource consumption<br>• effective storage capacity |
| | | | | **ALL-Healthy** | | |
| [44] | 2012 | Morphological | Shape and texture features | K-NN | 92.5 | • Using Marker-Controller Watershed algorithm in the segmentation step |
| [47] | 2014 | Shadowed C-Means clustering (SCM) | Shape, color, and texture features | EOC | 96.88 | • Using EOC Classifiers |
| [1] | 2018 | Histogram equalization thresholding | Shape, color, and texture features | SVM, K-NN, NB, and DT | 96.01 | • Appling three data normalization methods including z-score, min-max and grey scaling in the feature extraction step |
| [25] | 2019 | Zack algorithm and morphology | statistical and texture features | SVM and K-NN | 91.5 | • Providing a strong model against noisy datasets |
| | | | | **ALL(L1, L2, L3) –Healthy** | | |
| [45] | 2018 | Fuzzy C-Means clustering (FCM) | Shape and statistical | Multiple SVM | 97 | • Using FCM in the segmentation step for avoiding empty cluster |
| | | | | **AML-Healthy** | | |
| [3] | 2014 | K-Means | Hausdorff dimension, Shape, color, and texture | SVM | 98 | • Using feature extraction step with Local Binary Pattern (LBP) on the Hausdorff dimension |
| [74] | 2019 | Otsu's Thresholding | Morpholoyfeatures | RF | 89.6 | • Using morphological's characteristics in the feature extraction step<br>• Using SMOTE with a RF algorithm to overcome the imbalance of data |
| [17] | 2020 | Multi Otsu's Thresholding and morphology | Shape, size, and color features | RF | 93.45 | • Using Gini importance for optimum feature extraction and classification steps. |
| | | | | **ALL, AML-Healthy** | | |
| [66] | 2012 | – | Shape, size, and color features | K-NN | 86 | • Finding optimum value for K in classification step |
| [57] | 2017 | Otsu's Thresholding | Size, color, and texture features | GA-SVM | 99.5 | • Using Genetic algorithm and Radial Basis Kernel (RBF) in the classification step |

Table 2 (continued)

| Work | Year | Segmentation method | Feature Extraction method | Classifier(s) employed | Accuracy (%) | Remarks |
|---|---|---|---|---|---|---|
| [16] | 2020 | | Deep features | AlexNet-R and AlexNet-X WBC | 97.18 | • Presenting a method in term of classification technique and type of classification<br>• Simulation real environment with multi-database<br>• Using the method in the mobile devices |
| [20] | 2016 | Otsu's Thresholding | Nucleus region | NB | 80.88 | • Using otsu's thresholding in the segmentation step |
| [13] | 2019 | Watershed | Morphological, color, and texture features | RF | 95.86 | • Removing image background for identifying WBCs<br>• Using a binary mask for detecting damaged cells<br>• Generalizing the model by introducing new discriminative features |
| [64] | 2019 | CNN | Shape, texture, and complex features | SVM | 96.1 | • Using CNN in the feature extraction step<br>• Using PCA in the feature reduction step for generation optimum feature vectors |

Table 3 Deep learning classification methods of leukemia based works

| Work | Year | Segmentation method | Feature Extraction method | Classifier(s) employed | Accuracy (%) | Remarks |
|---|---|---|---|---|---|---|
| | | | | Leukemia (general)-Healthy | | |
| [21] | 2018 | Adaptive K-Means method (AKM) | Shape features | ANN, SVM and CNN | 93 | • Using AKM in the segmentation step for extracting the nucleus<br>• Implementing The model with ANN, CNN and SVM for finding the best classifier |
| [42] | 2020 | – | CNN features | LD, DT and K-NN | 100 | • Introducing traditional machine learning and deep learning with two leukemia models and comparing them |
| [72] | 2020 | – | Deep features | CNN based VGGNet-16 | 98.24 | • Optimum parameters and architecture<br>• More strong and accurate |
| | | | | ALL-Healthy | | |
| [68] | 2018 | | CNN features | FC | 96.6 | • Using the data augmentation to prevent overfitting |
| [4] | 2019 | – | CNN features | CNN | 88.25 | • Using the data augmentation step to avoid overfitting |
| [30] | 2019 | Mutual information-based hybrid model | Statistical and Local Directional Pattern(LDP) features | Deep learning | 98.7 | • Proposing new segmentation using the Mutual Information (MI)<br>• Introducing new classification using chronological Sine Cosine Algorithm (SCA) based on CNN |
| [15] | 2020 | – | Morphological and statistical features | SVM and ANN | 97.52 | • Using a new technique in the feature extraction step for splitting nucleus from cytoplasm features<br>• Using ANN and SVM in the classification step |
| | | | | ALL(L1, L2, L3) -Healthy | | |
| [58] | 2018 | Threshold | CNN features | CNN | 97.78 | • Using a new segmentation method to obtain better performance metrics |
| [62] | 2018 | Without segmentation | CNN features | FC | 99.5 | • Effective and no need for segmentation step<br>• Using the feature extraction with DCNN |
| [2] | 2019 | K-Medoid algorithm | Shape, visual and texture features | ANN | 98.6 | • Using the ranker search method for ranking and extracting the features<br>• Performing the segmentation step without need to crop the image explicitly<br>• Training the model with different illumination conditions for making a robust model |
| | | | | WBC | | |
| [75] | 2017 | – | CNN features | CNN | 88.5 | • Testing the model with blur images for making strong, effective against big invariance forms of distortion<br>• Using multi DNN in the classification step for achieving better performance metrics |
| [80] | 2017 | – | CNN features | FC | 92.8 | • Introducing new feature extraction method |
| [24] | 2018 | – | Hierarchical deep features | | 99.84 | |

Table 3 (continued)

| Work | Year | Segmentation method | Feature Extraction method | Classifier(s) employed | Accuracy (%) | Remarks |
|---|---|---|---|---|---|---|
| | | | | Inception and ResNet | | • using various Inception, ResNet and transfer learning in the classification step to prevent model overfitting and convergence problems<br>• Using Inception V3 for WBCs recognition |
| [40] | 2018 | Watershed | Color and Morphological features | CNN | 98.96 | • Removing the preprocessing step with using CNN<br>• Improving the feature extraction step with varying K-means clustering algorithm for extracting complex WBCs |
| [43] | 2018 | – | CNN features | Ensemble CNN | 96.63 | • Using Ensemble CNN in the classification step for achieving better performance metrics |
| [73] | 2018 | – | CNN features | Pattern Net fusion of Ensemble-CNN (PECNN) | 99.37 | • Using ensemble CNN and majority voting in the classification step<br>• Less computational cost and light weight<br>• Using the method in mobile applications |
| [11] | 2019 | | CNN features | FC | 90.39 | |
| [10] | 2020 | | CNN features | CNN | 96 | • Increasing model generalization with the type of segmentation<br>• Ineffective segmentation performance on the classification accuracy |

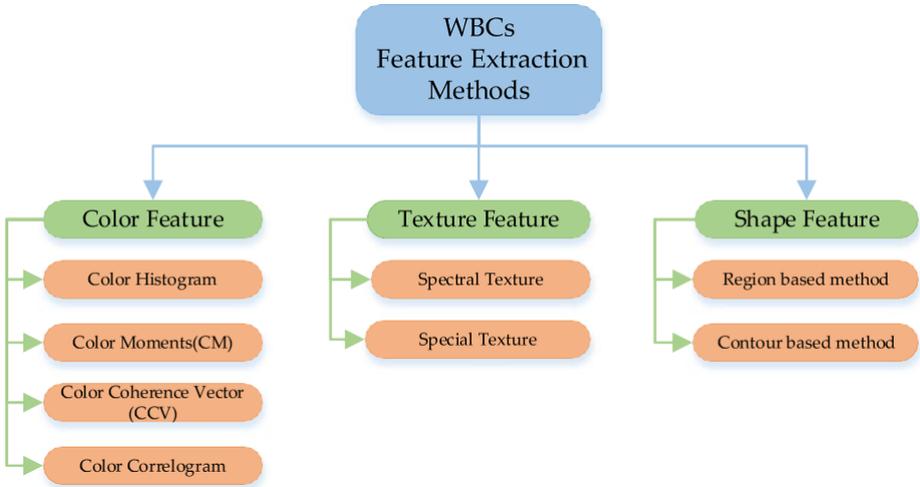

Fig. 10 WBCs feature extraction methods for Acute Leukemia and WBCs

Feature reduction is critical and sensitive in the classification task. In [5, 28, 32, 35, 45–47, 53, 61, 64, 70, 71], after the feature extraction stage, the feature selection step was performed.

### 4.6 Classification

Many methods can be used for the classification step. In this step, supervised and unsupervised algorithms are used for classification. We found out that almost works in the classification step used supervised algorithms on diagnosing acute leukemia and WBCs. Therefore, we have divided the classification step into three sections: traditional, DNN, and mixture methods. The base of this division was the type of supervised classification in the category. Figure 11 shows the taxonomy of all classification methods for acute leukemia and WBCs.

#### 4.6.1 Traditional methods for acute leukemia classification

In this section, we review a number of previous studies that have classified acute leukemia and WBCs with traditional networks. The properties extracted in these methods are performed manually and based on the characteristics of WBCs. A summary of previous works of acute leukemia detection and classification with traditional methods are shown in Table 2. Our purpose is to determine the relationship between the accuracy of the models and the classifier.

In traditional machine learning methods, the classification step is entirely dependent on the feature extraction step. In these methods, feature extraction is usually done manually. In [8], they used three CNN for features to be extracted automatically. With this technique, they automated the feature extraction step in the traditional model. Most models used the SVM Classifier, which had an accuracy of more than 96% [7, 23, 52, 62, 87]. Mirmohammadi model used to Multi SVM and Mohapatra employed ensemble classifier in their model, one of which was an SVM. The accuracy of these models was more than 96%. As we have seen, SVM Classifier has been used more in traditional machine learning because achieved with it has been good.

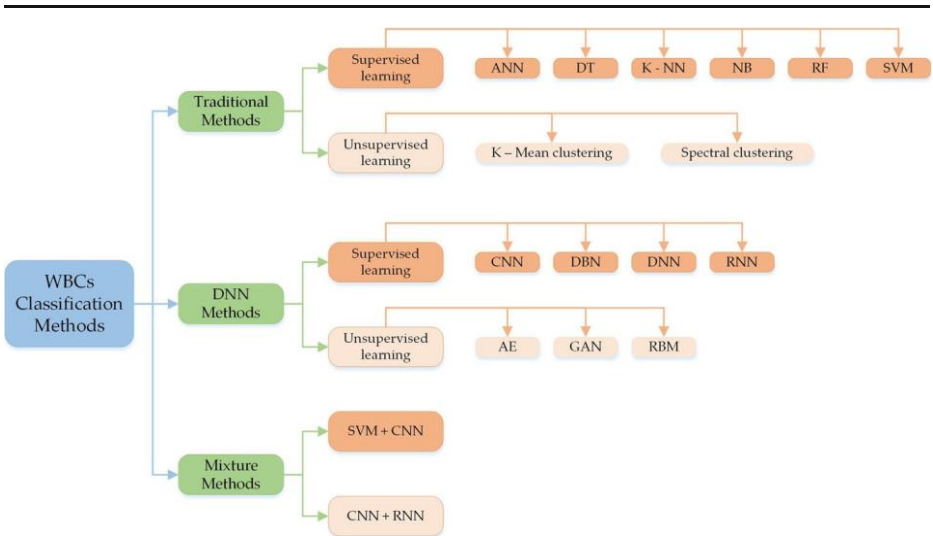

Fig. 11 Taxonomy of classification methods for acute leukemia and WBCs

### 4.6.2 Deep-learning-based methods for acute leukemia classification

In this section, we have shown several previous works that have classified acute leukemia and WBCs with DNN. Table 3 shows a summary of previous leukemia detection and classification works with DNN methods.

In this section, we analyze previous work that has used DNN methods to diagnose and classify acute blood leukemia. Traditional models used to manually feature extraction step, but this step is automatically in DNN methods. Therefore, researchers prefer to use DNN methods to identify and classify acute leukemia. The automatic feature extraction step allows the DNN to work end-to-end and eliminates the cost of feature extraction and reduction steps. These models can be categorized into different types such as Deep Belief Network (DBN), CNN, LSTM, RNN, Deep Autoencoders (AE), Generative Adversarial Network (GAN), Restricted Boltzmann Machine (RBM) and etc. In the following, we consider the previous work of the classification of acute leukemia and WBCs by DNN:

Many DNN models used CNN to classify acute leukemia. In [4, 13, 20, 48, 66, 70, 77], CNN was used for classification, and the accuracy of all of them was more than 96%. In some models such as [15, 17], ensemble CNN was used for classification, and the accuracy of the models was over 99%. Some studies have also proposed a new approach for acute blood leukemia classification by modifying classical CNN [1, 66]. Therefore, we can conclude from the study of DNN models for acute leukemia categorization that CNN is used more than other methods, and the reason is the high efficiency of this model.

### 4.6.3 Mixed methods for acute leukemia classification

In this section, the methods that have used two or more methods in the classification step of healthy cells and acute leukemia (see Table 4) are reviewed. Like DNN methods, the accuracy of these methods is high. Therefore, the use of these methods has established consideration in current years. As we can see, the combination of SVM and CNN has been used more.

Table 4 Mixture classification methods of leukemia based works

| Work | Year | Segmentation method | Feature Extraction method | Classifier(s) employed | Accuracy (%) | Remarks |
|---|---|---|---|---|---|---|
| [70] | 2017 | – | DCNN | Leukemia (general)-Healthy SVM+ MLP+ RF | 100 | • Using CNN for automatic feature extraction<br>• Using EOC (SVM, MLP, and RF) for getting optimum classification |
| [41] | 2019 | | ResNet features | Two Inception ResNet WBC | 90 | • Using transfer learning with Inception and ResNets architecture<br>• Employing the data augmentation for increasing training samples |
| [29] | 2017 | Countor Aware based on Fully Convolutional Network | CNN features | FRCN+ ELM | 95.1 | • Using a new segmentation method (Contour Aware segmentation with Fully CNN)<br>• Using CNN based on ELM in the classification step |
| [39] | 2018 | – | Deep features | CNN+ RNN | 90.79 | • Using Xception and RNN model for training big medical data<br>• Appling transfer learning method to fast train, converge and more accurate weight parameters |
| [26] | 2019 | Thresholding and morphology | Shape, color, and texture features | SVM+NN | 98.8 | • Using a robust method for WBCs segmentation<br>• Generating mask for discovering WBCs from sub-images<br>• Dividing classification task by two SVM (SVM1 and SVM2) and NN for decreasing classification time |

In this section, we analyze the past method on the acute leukemia detection and classification that has used a mixture of traditional and DNN methods in the classification. Our observations show that researchers have focused less on mixture classification methods. Therefore, the number of studies of these methods is much less than the previous two methods. We describe the previous work of the combined methods of acute blood leukemia:

We analyze the performance of acute leukemia classification based on three types of classifiers, including traditional, DNN, and mixed, in this section. In most traditional network methods, the SVM classifier is used more for classification, which also has acceptable accuracy. In many cases, DNN were more accurate than traditional networks. Among DNN works, CNN were the most widely used. In the mixture work, the two classifiers SVM and CNN were used more and had good results. In addition, in recent years, most ALL detection and classification tasks are based on DNN. Among the DNN, CNN have received special attention and have achieved very good results.

## 5 Metrics for performance evaluation

Models performance evaluation is needed in entirely automatic illness detection, and classification methods are calculated from a confusion matrix. In this study, the confusion matrix denotes the differences in opinion between the classifier and the hematologist (see Table 5). Positive and negative in a two states classification in order are as recognized and rejected. So, TPR (True Positive Rate or correctly identified) defined as Sensitivity Rate, TNR (True Negative Rate or incorrectly identified) defined as Recall or Specificity Rate, FPR (False Positive Rate or correctly rejected), and FNR (False Negative Rate or mistakenly rejected) can be defined in confusion matrix for classifier performance evaluation [43] [32]. Evaluation metrics, including measures and formulas, are shown in Table 6.

In the following, we analyze the previous studies that reported the above metric in Tables 7, 8, and 9. Comparison between models for Leukemia (General) classification in terms of evaluation metrics is shown in Table 7. The best evaluation metric of Table 7 marked in bold font.

Comparison between models for ALL classification in terms of evaluation metrics is shown in Table 8. Among these methods, the work that marked in bold font has better efficiency metrics.

Comparison between models for classification types of WBC in terms of evaluation metrics is shown in Table 9. Most models in this table have close performance.

## 6 Discussion

We analyzed different steps for diagnosing and classifying acute leukemia and WBCs. The classification step performs an essential role in model evaluation metrics. Thus, we focused on the type of classifier in the classification step. In this section, we use studies that have used both traditional and DNN methods in the classification step for acute leukemia with similar previous steps. Our goal is to compare the performance of the categories between traditional and DNN methods.

Table 5 Confusion Matrix for classifier performance evaluation of acute leukemia

|  |  | Hematologist Opinion | |
|---|---|---|---|
|  |  | Positive (Leukemia) | Negative (Healthy) |
| Classifier Prediction | Positive (Leukemia) | TPR | FPR |
|  | Negative (Healthy) | FNR | TNR |

| Table 6  Evaluation metrics | Formula (× 100) |
|---|---|
| Precision | $\frac{TPR}{TPR + FPR}$ |
| Recall (Sensitivity) | $\frac{TPR}{TPR + FNR}$ |
| Specificity | $\frac{TNR}{TNR + FPR}$ |
| Accuracy | $\frac{TPR + TNR}{TPR + TNR + FP + FN}$ |
| F-Measure | $2 \times \frac{Precision \times Recall}{Precision + Recall}$ |

Vogado et al. used a traditional model to diagnose and classify acute leukemia in healthy and unhealthy images [70]. The model used the CNN for the feature extraction step and the EOC (SVM, MLP and RF) in the classification step. The model had achieved 100% accuracy. Vagado et al. used the same model [70] in the Model [71], and the main difference was in the classification step. They used the SVM, MLP, RF, K-NN classifiers in the Model [71]. Each time the model was trained and tested with one of these classifiers. The best accuracy among these models was achieved by SVM with 100%. They also trained and evaluated the model once with three CNN (AlexNet, CaffeNet and vgg-f) separately. Performance metrics (Accuracy, Precision and Recall) of vgg-f were better than the others with 99%. They also trained and tested the model by combining these three classifiers, which did not improve the performance metrics. By comparing the models performed in [70, 71], we can be concluded that traditional models are more efficient than DNN models in diagnosing and classifying acute leukemia and WBCs. Is this statement always correct? For further investigation, we analyze the work similar to the Vogado models.

Loey et al. used two completely like models to classify normal and abnormal acute blood leukemia images [42]. In the first model, they used the traditional model and in the second model employed DNN. The feature extraction step of both models was done by CNN (AlexNet) network. The difference between these models was in the classification step. In the traditional model, SVM, LD, DT, and K-NN were used in the DNN model, and they used AlexNet. In the traditional model, SVM with performance metrics including Precision, Recall, Accuracy and Specificity above 99% (the best classifier), and in the DNN model were 100% (See Table 7). In this paper, we can have a good understanding of the classification step in the traditional model. In the traditional model, each time model is trained and tested with one of the above algorithms. We see that SVM has had the best results. Also, by comparing these two models, it can be inferred that DNN methods usually achieve better efficiency metrics than traditional methods. Now, we ask a question: Are the DNN models always better than the traditional models and get good results? According to Loey model, the answer to this question is no.

Table 7  Comparison between previous models for Leukemia (General) classification

| Work | Year | Classifier employed | P (%) | R (%) | S (%) | A (%) | F1-score (%) |
|---|---|---|---|---|---|---|---|
| [71] | 2018 | SVM | 99.2 | 99.2 | – | 99.2 | – |
| [21] | 2018 | SVM, CNN | 100 | 88.34 | 100 | 93 | – |
| [41] | 2019 | Two Inception ResNet | 84 | 85 | – | 90 | 84 |
| [61] | 2020 | SVM | 93.43 | 99.55 | 92.92 | 96.11 | 96.22 |
| [26] | 2020 | CNN (AlexNet) | 100 | 100 | 100 | 100 | – |

Table 8 Comparison between previous models for ALL

| Work | Year | Classifier employed | P (%) | R (%) | S (%) | A (%) | F1-score (%) |
|---|---|---|---|---|---|---|---|
| [47] | 2014 | EOC | – | 94.93 | 95 | 96.88 | – |
| [25] | 2019 | SVM-R | – | 90 | 92 | 91.5 | – |
| [36] | 2020 | ANN | – | 100 | 95.31 | 97.52 | 97.44 |

We studied different traditional, DNN and mixed methods to categorize WBCs in blood smear images. After reviewing previous studies, we found that the number of traditional and DNN approaches are equally (See Fig. 5). Previous research in the diagnosis and category of acute leukemia were more traditional, but recent works on this subject have more used DNN. SVM in traditional networks and CNN in DNN are more had been used. The reason for the more use of these two classifiers is to obtain good performance metrics in the results on the test databases.

Table 9 Comparison between previous works for classification the types of WBC

| Work | Year | WBC cell types | P (%) | R (%) | S (%) | A (%) | F1-score (%) |
|---|---|---|---|---|---|---|---|
| [75] | 2017 | B | – | 100 | 100 | 100 | – |
| | | E | – | 67 | 100 | 99 | – |
| | | L | – | 61 | 99 | 95 | – |
| | | M | – | 32 | 100 | 96 | – |
| | | N | – | 93 | 100 | 95 | – |
| [43] | 2018 | B | – | 100 | 100 | 100 | – |
| | | E | – | 100 | 100 | 100 | – |
| | | L | – | 95.04 | 97.3 | 95.51 | – |
| | | M | – | 75 | 91.95 | 91.57 | – |
| | | N | – | 75.86 | 100 | 96.07 | – |
| [39] | 2018 | B | – | – | – | – | – |
| | | E | 93 | 91 | 97.86 | 96.24 | 92 |
| | | L | 100 | 100 | 99.94 | 99.92 | 100 |
| | | M | 96 | 80 | 98.96 | 94.44 | 88 |
| | | N | 78 | 92 | 90.88 | 91.08 | 84 |
| [50] | 2018 | B | – | – | – | – | – |
| | | E | 94 | 84 | 98.28 | 94.8 | 89 |
| | | L | 100 | 100 | 100 | 100 | 100 |
| | | M | 100 | 75 | 100 | 93.86 | 86 |
| | | N | 71 | 95 | 86.37 | 88.66 | 82 |
| [26] | 2019 | B | – | – | – | – | – |
| | | E | – | 98.7 | 99.8 | 99.8 | 99.7 |
| | | L | – | 100 | 100 | 100 | 100 |
| | | M | – | 100 | 99.8 | 99.8 | 99.1 |
| | | N | – | 98.6 | 98 | 99.5 | 99 |
| [11] | 2019 | B | – | – | – | – | – |
| | | E | 96 | 84 | 98.87 | 95.05 | 89 |
| | | L | 99 | 100 | 99.78 | 99.84 | 100 |
| | | M | 100 | 81 | 100 | 95.33 | 90 |
| | | N | 74 | 97 | 88.51 | 90.55 | 84 |
| [10] | 2020 | B | – | – | – | – | – |
| | | E | 93 | 93 | 97.5 | 96.48 | 93 |
| | | L | 100 | 100 | 99.89 | 99.92 | 100 |
| | | M | 99 | 99 | 99.58 | 99.45 | 99 |
| | | N | 93 | 92 | 97.5 | 95.93 | 92 |

If we want to have a general conclusion, it seems that the results of model evaluation are highly dependent on the classification step in the acute leukemia identification and categorization. But it does not mean that the role of other steps does not matter. The role of other complementary steps determines the performance of the classifier and thus the efficiency of the model.

# 7 Conclusions

Cancer is one of the greatest illnesses in the world today. Leukemia is commonly in almost all ages and separated into acute and chronic types. Acute leukemia is a dangerous type that is more likely to die. In the paper, we first comprehensively review the steps in the identification and classification of acute leukemia and WBCs based computer system. The steps for diagnosing and classifying acute leukemia and WBCs include data augmentation, image preprocessing, segmentation, feature extraction, feature selection, and classification. Then, we analyzed previous work on this subject by the focus on the classification step. We have divided the previous classification methods into three categories, including traditional, DNN, and hybrid based on the type of classifier and compared the accuracy of each of them. We realized that SVM among the traditional methods and CNN among DNN methods have attained the highest accuracy. The combination of these two classifiers had achieved good accuracy in mixture methods, too. In classification issues, the accuracy of the model is very important, but not enough. For this reason, we evaluated previous works that reported other terms of performance metrics. With this comparison, we found that the works in which the CNN classifier was used for classification have higher performance metrics than other methods. Therefore, we can be concluded that the models use different CNN methods to diagnose and classify acute leukemia and WBCs are more accurate and efficient than the rest. Of course, the combination of CNN classifiers can also achieve perfect results.

For future work, we propose to use the new generation of CNN for the classification of acute leukemia and WBCs to achieve good results in this subject. We recommend that CNN with SVM classifiers mixture employed in classification step for achieving the best performance metrics in identification and classification of acute leukemia and WBCs, too. It is also better to pay more attention to the steps before classification in future research so that acute leukemia can be diagnosed and classified in less time and faster by reducing the complexity and dimensions of the features.

# References

1. Abdeldaim, A.M.; Sahlol, A.T.; Elhoseny, M.; Hassanien, A. E, "computer-aided acute lymphoblastic leukemia diagnosis system based on image analysis," in *advances in soft computing and machine learning in image processing*, Berlin/Heidelberg, Germany, 2018.
2. Acharya V, Kumar P (2019) Detection of acute lymphoblastic leukemia using image segmentation and data mining algorithms. Med Biol Eng Comput 57:1783–1811
3. Agaian, S.; Madhukar, M.; Chronopoulos, A. T, "A new acute leukemia automated classification system," Comp Methods Biomech Biomed Eng: Imaging Visual, vol. 6, no. 3, p. 303–314, 2018.

4. Ahmed N, Yigit A, Isik Z, Alpkocak A (2019) Identification of leukemia subtypes from microscopic images using Convolutional Neural Network. MDPI (Diagnostics) 9(3):104
5. Al-jaboriy SS, Sjarif NNA, Chuprat S, Abduallah WM (2019) Acute lymphoblastic leukemia segmentation using local pixel information. Pattern Recogn Lett 125:85–90
6. Al-jaboriy, S.; Sjarif, N.; Chuprat, S, "Segmentation and detection of acute leukemia using image processing and machine learning techniques: a review," p. 511–531, 2019.
7. Alsalem MA, Zaidan AA, Zaidan BB, Hashim M, Madhloom HT, Azeez ND, Alsyisuf S (2018) A review of the automated detection and classification of acute leukaemia: coherent taxonomy, datasets, validation and performance measurements, motivation, open challenges and recommendations. Comput Methods Prog Biomed 158:93–112
8. Anilkumar, K.K.; Manoj, V.J.; Sagi, T. M, "colour based image segmentation for automated detection of Leukaemia: a comparison between CIELAB and CMYK colour spaces," in *international conference on circuits and Systems in Digital Enterprise Technology (ICCSDET)*, Kottayam, India, 2018.
9. Bagasjvara, R.G.; Candradewi, I.; Hartati, S.; Harjoko, a, "automated detection and classification techniques of acute leukemia using image processing: a review," in *2nd international conference on science and technology computer (ICST)*, Yogyakarta, Indonesia, 2016.
10. Banik PP, Saha R (2020) Kim, K, "an automatic nucleus segmentation and CNN model based classification method of white blood cell,". Expert Syst Appl 149:113211
11. Banik, P.P.; Saha, R.; Kim, K-D, "fused convolutional neural network for white blood cell image classification," in *international conference on artificial intelligence in information and communication (ICAIIC)*, Okinawa, Japan, 2019.
12. Bennett JM, Catovsky D, Daniel MT, Flandrin G, Galton DAG, Gralnick HR, Sultan C (1976) Proposals for the classification of the acute leukaemias. French-American-British (FAB) co-operative group. Brit J Hematol 33(4):451–458
13. Benomar, M.L.; Chikh, A.; Descombes, X.; Benazzouz, M, "Multi features based approach for white blood cells segmentation and classification in peripheral blood and bone marrow images Int J Biomed Eng Technol,* 2019.
14. Bibi N, Sikandar M, Din IU, Almogren A, Ali S (2020) IoMT-based automated detection and classification of leukemia using deep learning. J Healthcare Eng 2020:1–12
15. Bodzas A, Kodytek P (2020) Zidek, J, "automated detection of acute lymphoblastic leukemia from microscopic images based on human visual perception,". Front Bioeng BiotechnolExpert 8:1005
16. Claro, M.; Vogado, L.; Veras, R.; Santana, A.; Tavares, J.; Santos, J.; Machado, V, "convolution neural network models for acute leukemia diagnosis," in *international conference on systems,* Signals and Image Processing (IWSSIP), Niteroi, Brazil, 2020.
17. Dasariraju S, Huo M, McCalla S (2020) Detection and classification of immature leukocytes for diagnosis of acute myeloid leukemia using Random Forest algorithm. *MDPI (Bioengineering)* 7(4):120
18. Dhanachandra N, Manglem K, Jina Chanu Y (2015) Image segmentation using K-means clustering algorithm and subtractive clustering algorithm. Procedia Comp Sci 54:764–771
19. Fisher RA (1936) The use of multiple measurements in taxonomic problems. Hum Genet 7(2):179–188
20. Gautam, A.; Singh, P.; Raman, B.; Bhadauria, H, "automatic classification of leukocytes using morphological features and Naïve Bayes classifier," in *IEEE region 10 conference (TENCON)*, Singapore, 2016.
21. Gayathri S, Jyothi RL (2018) An automated leucocyte classification for leukemia detection. Int 744 Res J Eng Technol (IRJET) 5(5):4254–4264
22. Ghane N, Vard A, Talebi A, Nematollahy P (2019) Classification of chronic myeloid leukemia cell subtypes based on microscopic image analysis. Med EXCLI 18:382–404
23. Gonzalez RC, Woods RE (2018) Digital image processing, New York, USA: Pearson, 330 Hudson street. New York, NY 10013
24. Habibzadeh, M.; Jannesari, M.; Rezaei, Z.; Baharvand, H.; Totonchi, M, "automatic white blood cell classification using pre-trained deep learning models: Resnet and inception," in *tenth international conference on machine vision (ICMV)*, Vienna, Austria, 2018.
25. Hariprasath, S.; Dharani, T.; Mohammad, S.; Bilal, N, "automated detection of acute lymphocytic leukemia using blast cell morphological features," in *2nd international conference on advances in science and technology (ICAST)*, Mumbai, India, 2019.
26. Hegde RB, Prasad K, Hebbar H, Sing BMK, Sandhya I (2019) Automated decision support system for detection of leukemia from peripheral blood smear images. Digital Imaging 33:361–374
27. Hosseinzadeh Kassani, S.; Hosseinzadeh Kassani, P.; Wesolowski, M.J.; Schneider, K.A.; Deters, R. A, "A hybrid deep learning architecture for leukemic B-lymphoblast classification," in *International Conference on Information and Communication Technology Convergence (ICTC)*, Jeju, Korea (South), 2019.
28. Huang D-C, Hung K-D, Chan Y-K (2012) A computer assisted method for leukocyte nucleus segmentation and recognition in blood smear images. J Syst Softw 85(9):2104–2118

29. Imran Razzak, M.I; Naz, S, "Microscopic blood smear segmentation and classification using deep contour aware CNN and extreme machine learning," in *IEEE conference on computer vision and pattern recognition workshops (CVPRW)*, Honolulu, HI, USA, 2017.
30. Jha KK, Dutta HS (2019) Mutual information based hybrid model and deep learning for acute lymphocytic leukemia detection in single cell blood smear images. C*omp Methods Programs Biomed* 179:104987
31. Jin, X.; Han, J. K-Medoids Clustering Ed.; Sammut, C, Webb G. I, Encyclopedia of machine learning, Boston: Springer, Boston, MA, 2016.
32. Jothi G, Inbarani HH, Azar AT, Devi KR (2019) Rough set theory with jaya optimization for acute lymphoblastic leukemia classification. Neural Comput & Applic 31:5175–5194
33. Kass, M.; Witkin, A.; Terzopoulos, D. Snakes, "active contour modelsn," .*International Journal of Computer Visio,* p. 321–331, 1988.
34. Krizhevsky, A.; Sutskever, I.; Hinton, G. E, "ImageNet classification with deep convolutional," Adv Neural Inform Process Syst (NIPS)*,* vol. 25, pp. 1097–1105, 2012.
35. Kumar, P.; Udwadia, S.N., "automatic detection of acute myeloid leukemia for microscopic blood smear image," in *international conference on advances in computing,* Communications and Informatics (ICACCI), Udupi, India, 2017.
36. Labati R.D.; Piuri, V.; Scotti, F, "All-IDB: the acute lymphoblastic leukemia image database for image processing," in *18th IEEE International Conference on Image Processing*, Brussels, Belgium, 2018.
37. Lai JZC, Juan EYT, Lai FJC (2013) Rough clustering using generalized fuzzy clustering algorithm. Pattern Recogn 46(9):2538–2547
38. Laosai J, Chamnongthai K (2018) Classification of acute leukemia using medical knowledge-based morphology and cd marker. Biomed Signal Process Control 44:127–137
39. Liang G, Hong H, Xie W, Zheng L (2018) Combining convolutional neural network with recursive neural network for blood cell image classification. IEEE Access 6:36188–36197
40. Lin L, Wang W, Chen B (2018) Leukocyte recognition with convolutional neural network. Algorithms Comput Technol 13:1–8
41. Liu, Y. and Long, F, "Acute lymphoblastic leukemia cells image analysis with deep bagging ensemble learning," in *CNMC challenge: classification in Cancer cell imaging*, Springer, Singapore, 2019.
42. Loey M, Naman M, Zayed H (2020) Deep Transfer Learning in diagnosing leukemia in blood cells. *MDPI (Computers)* 9(3):29
43. Macawile, M.J.; Quiñones, V.V.; Ballado, A.; Cruz, J.D.; Caya, M. V, "white blood cell classification and counting using convolutional neural network," in *3rd international conference on control and robotics engineering (ICCRE)*, Nagoya, Japan, 2018.
44. Madhloom HT, Kareem SA (2012) Ariffin, H, "a robust feature extraction and selection method for the recognition of lymphocytes versus acute lymphoblastic leukemia," in *advanced computer science applications and technologies (ACSAT).* Kuala Lumpur, Malaysia
45. Mirmohammadi P, Rasooli A, Ashtiyani M, Moradi Amin M (2018) Automatic recognition of acute lymphoblastic leukemia using multi-SVM classifier. Biology 115:1512
46. Mishra S, Majhi B, Sa PK (2019) Texture feature based classification on microscopic blood smear for acute lymphoblastic leukemia detection. *Biomedical Signal Processing and Control* 47:303–311
47. Mohapatra S, Patra D (2014) N ensemble classifier system for early diagnosis of acute lymphoblastic leukemia in blood microscopic images. Neural Comput & Applic 24:1887–1904
48. Otsu N (1979) A threshold selection method from gray-level histograms. IEEE Trans Syst 9(1):62–66
49. Pandey P, Pallavi S, Pandey SC (2019) Pragmatic Medical Image Analysis and Deep Learning: An Emerging Trend. Advanc Mach Intell Interactive Med Image Analy:1–18
50. Pang S, Du A, Orgun MA, Yu Z (2019) A novel fused convolutional neural network for biomedical image classification. Med Biol Eng Comput 57:107–121
51. Pansombut, T.; Wikaisuksakul, S.; Khongkraphan, K.; Phon-on, a, "convolutional neural networks for recognition of lymphoblast cell images," Comput Intell Neurosci*,* 2019.
52. Patel N, Mishra A (2015) Automated leukemia detection using microscopic images. Procedia Comput Sci 58:635–642
53. Patil S, Rathod PP, Patane S, Patil M (2020) Acute lymphoblastic leukemia detection in human blood using microscopic image. Int J Future Gen Comm Networking 13:1539–1544
54. Pearson K (1901) On lines and planes of closest fit to systems of points in space. Philos Mag 2(1):559–572
55. Putzu, L.; Di Ruberto, C, "white blood cells identification and classification from leukemic blood image," in *international work-conference on bioinformatics and biomedical engineering (IWBBIO)*, Granada, Spain, 2013.
56. Ravikumar S (2015) Image segmentation and classification of white blood cells with the extreme learning machine and the fast relevance vector machine. Artificial Cells, Nanomed Biotechnol 44(3):985–989

57. Rawat, J.; Singh, A.; HS, B.; Virmani, J.; Devgun, J. S, "Computer assisted classification framework for prediction of acute lymphoblastic and acute myeloblastic leukemia," Biocybernetics Biomed Eng, vol. 37, no. 4, p. 637–654, 2017.
58. Rehman A, Abbas N, Saba T, Ur-Rahman SI, Mehmood Z, Kolivand H (2018) Classification of acute lymphoblastic leukemia using deep learning. Microsc Res Tech 81(11):1310–1317
59. Safuan SNM, Tomari MRM, Zakaria WNW, Sing BMK, Mohd MNH, Suriani NS (2019) Computer aided system for lymphoblast classification to detectacute lymphoblastic leukemia. Indonesian J Electrical Eng Comp Sci 14:597–607
60. Sah, S, "Machine Learning: A Review of Learning Types," *Preprints,* 2020.
61. Sahlol, A.T.; Kollmannsberger, P.; Ewees, A. A, "Efficient classification of white blood cell leukemia with improved swarm optimization of deep features," Sci Rep, vol. 10, p. 2536, 2020.
62. Shafique S, Tehsin S (2018) Acute lymphoblastic leukemia detection and classification of its subtypes using pretrained deep convolutional neural networks. Technol Cancer Res Treatment 17:1–7
63. Shafique, S.; Tehsin, S, "Computer-aided diagnosis of acute lymphoblastic leukaemia," *Computational and Mathematical Methods in Medicine,* p. 6125289, 2018.
64. Shahin AI, Guo Y, Amin KM, Sharawi AA (2019) White blood cells identification system based on convolutional deep neural learning networks. Comput Methods Prog Biomed 168:69–80
65. Singhal, V.; Singhal, P, "local binary pattern for automatic detection of acute lymphoblastic leukemia," in *twentieth National Conference on communications (NCC)*, Kanpur, India, 2014.
66. Supardi, N.Z.; Mashor, M.Y.; Harun, N.H.; Bakri, A.; Hassan, R, "classification of blasts in acute leukemia blood samples using K-nearest neighbor," in *IEEE 8th international colloquium on signal processing and its applications*, Malacca, Malaysia, 2012.
67. Terwilliger T, Abdul-Hay M (2017) Acute lymphoblastic leukemia: a comprehensive review and 2017 update. Blood Cancer J 7:e577
68. Thanh TTP, Vununu C, Atoev S, Lee S-H, Kwon K-R (2018) Leukemia blood cell image classification using convolutional neural network. Int J Comp Theory Eng 10(2):54–58
69. Umamaheswari D, Geetha S (2018) A framework for efficient recognition and classification of acute lymphoblastic leukemia with a novel customized-KNN classifier. Comp Inform Technol (CIT) 26:131–140
70. Vogado, L.H.S.; Veras, R., De M.S.; Andrade, A.R.; De Araujo, F.H.D.; Silva, R.R.V.; Aires, K.R.T, "Diagnosing leukemia in blood smear images using an ensemble of classifiers and pre-trained Convolutional Neural Networks," in *30th SIBGRAPI Conference on Graphics, Patterns and Images (SIBGRAPI)*, Niteroi, Brazil, 2017.
71. Vogado LH, Veras RM, Araujo FH, Silva RR, Aires KR (2018) Leukemia diagnosis in blood slides using transfer learning in CNNs and SVM for classification. Eng Appl Artif Intell, 72:415–422
72. Vogado, L.H; Veras, R.M.; Aires, KR, ""LeukNet" - a model of convolutional neural network for the diagnosis of leukemia," in *ANAIS ESTENDIDOS DA conference on graphics,* Patterns and Images (SIBGRAPI), Porto Alegre, Brasileira, 2020.
73. Wang JL, Li AY, Huang M, Ibrahim AK, Zhuang H, Ali AM (2018) "classification of white blood cells with PatternNet-fused Ensemble of Convolutional Neural Networks (PECNN)," in *IEEE international symposium on signal processing and information technology (ISSPIT)*. Louisville, KY, USA
74. Wiharto, W.; Suryani, E.; Putra, Y. R, "Classification of blast cell type on AML based on image morphology of white blood cells," Telecomm Computing Electronics Control (TELKOMNIKA), vol. 17, p. 645–652, 2019.
75. Yu, W.; Chang, J.; Yang, C.; Zhang, L.; Shen, H.; Xia, Y.; Sha, J. "automatic classification of leukocytes using deep neural network," in *in proceedings of the 2017 IEEE 12th international conference on ASIC (ASICON)*, Guiyang, China, 2017.
76. Zack GW, Rogers WE, Latt SA (1977) Automatic measurement of sister chromatid exchange frequency. J Histochem Cytochem 25:741–753
77. Zadeh LA (1965) Fuzzy sets. Inf Control 8(3):338–353
78. Zaitoun NM, Aqel MJ (2015) Survey on image segmentation techniques. Procedia Comp Sci 65:797–806
79. Zhang Q, Xie Q, Wang G (2016) A survey on rough set theory and its applications. AAI Trans Intell Technol 1(4):323–333
80. Zhao J, Zhang M, Zhou Z, Chu J, Cao F (2017) Automatic detection and classification of leukocytes using convolutional neural networks. Med Biol Eng Comput 55:1287–1301
81. Pardakhti N, Sajedi H (2020) Brain age estimation based on 3D MRI images using 3D-convolutional neural network. Multimed Tools Appl 79(33–34):25051–25065

Affiliations

Mohammad Zolfaghari[1] · Hedieh Sajedi[2]

✱ Hedieh Sajedi
hhsajedi@ut.ac.ir

   Mohammad Zolfaghari
mmzolfaghari@ut.ac.ir

[1]  Department of Computer Science, University of Tehran, Kish International Campus, Kish, Iran

[2]  Department of Mathematics, Statistics and Computer Science, University of Tehran, Tehran, Iran